\documentclass[10pt,twocolumn,letterpaper]{article}

\usepackage{cvpr}              

\usepackage{graphicx}
\usepackage{amsmath}
\usepackage{amssymb}
\usepackage{booktabs}
\usepackage{multirow}
\usepackage{booktabs}
\usepackage{caption}

\usepackage{algorithm}
\usepackage{algpseudocode}

\usepackage[pagebackref,breaklinks,colorlinks]{hyperref}

\usepackage[capitalize]{cleveref}
\crefname{section}{Sec.}{Secs.}
\Crefname{section}{Section}{Sections}
\Crefname{table}{Table}{Tables}
\crefname{table}{Tab.}{Tabs.}

\def\cvprPaperID{9933} 
\def\confName{CVPR}
\def\confYear{2023}

\begin{document}

\title{TMO: Textured Mesh Acquisition of Objects with a Mobile Device by using Differentiable Rendering}

\author{Jaehoon Choi\:\textsuperscript{\rm 1,2} \:\: Dongki Jung\:\textsuperscript{\rm 1} \:\: Taejae Lee\:\textsuperscript{\rm 1} \:\: Sangwook Kim\:\textsuperscript{\rm 1} \:\: Youngdong Jung\:\textsuperscript{\rm 1} \\ Dinesh Manocha\:\textsuperscript{\rm 2} \:\:Donghwan Lee\:\textsuperscript{\rm 1}\\
$^{1}$NAVER LABS \:\: $^{2}$University of Maryland}

\maketitle

\begin{abstract}
    We present a new pipeline for acquiring a textured mesh in the wild with a single smartphone which offers access to images, depth maps, and valid poses. 
    Our method first introduces an RGBD-aided structure from motion, which can yield filtered depth maps and refines camera poses guided by corresponding depth. 
    Then, we adopt the neural implicit surface reconstruction method, which allows for high-quality mesh and develops a new training process for applying a regularization provided by classical multi-view stereo methods.   
    Moreover, we apply a differentiable rendering to fine-tune incomplete texture maps and generate textures which are perceptually closer to the original scene. 
    Our pipeline can be applied to any common objects in the real world without the need for either in-the-lab environments or accurate mask images.    
    We demonstrate results of captured objects with complex shapes and validate our method numerically against existing 3D reconstruction and texture mapping methods. 
\end{abstract}

\section{Introduction}
\label{sec:intro}
Recovering the 3D geometry of objects and scenes is a longstanding challenge in computer vision and is essential to a broad range of applications. Depth sensing technologies range from highly specialized and expensive turn-table 3D scanners and structured-light scanners to commodity depth sensors. More recently, advances in mobile devices have developed a new method for 3D capture of real-world environments with high-resolution imaging and miniaturized LiDAR. Specifically, the modern smartphone such as iPhone 13 Pro are equipped with RGB camera, accelerometer, gyroscope, magnetometer, and LiDAR scanner. These various sensors can provide high-resolution images, low-resolution depth from the LiDAR scanner, and associated camera poses offered by off-the-shelf visual-inertial odometry (VIO) systems such as ARKit \cite{arkit} and ARCore \cite{ARCore}.



\begin{figure}[t]
  \centering
    \includegraphics[width=0.99\linewidth]{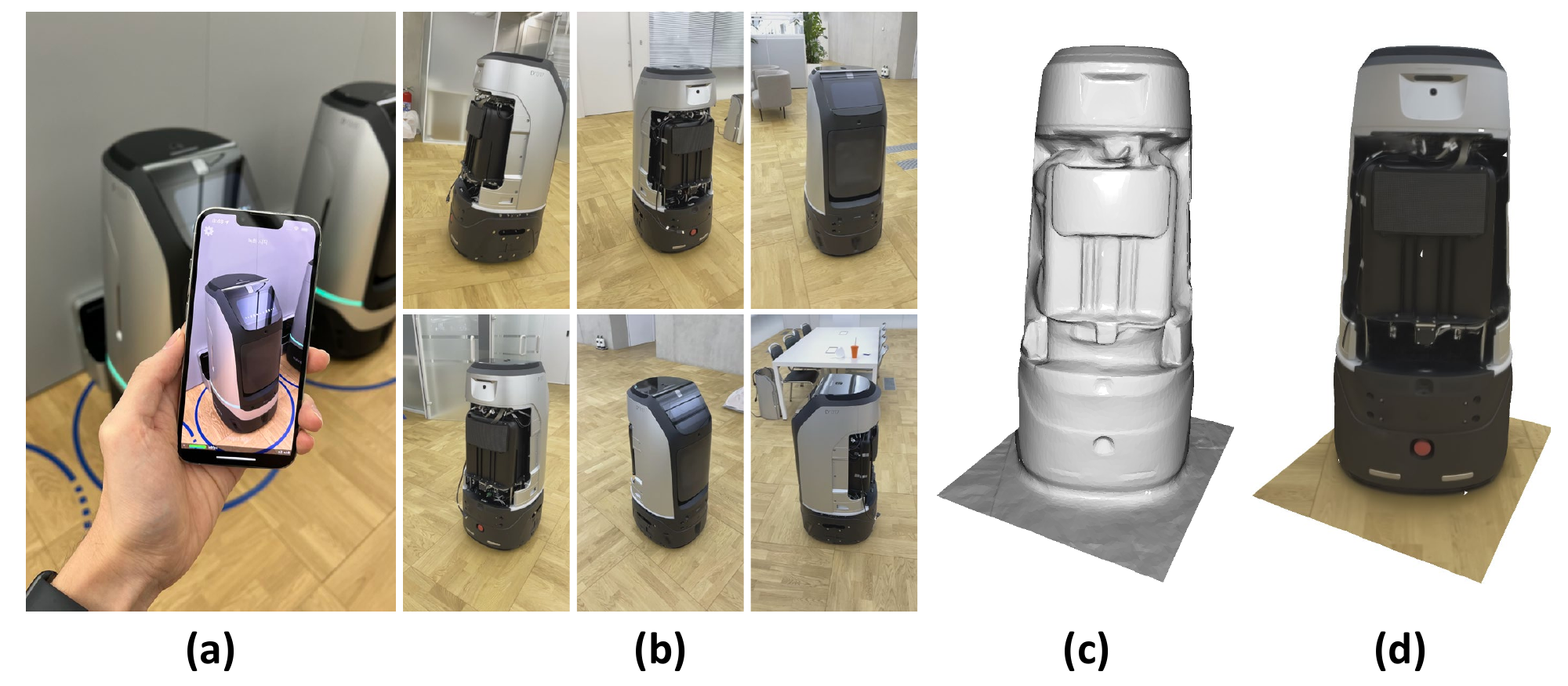}
    \vspace{-3mm}
   \caption{ Example reconstruction results collected from a smartphone in the wild. (a) Data acquisition setup. (b) Images captured from a smartphone. (c) A reconstructed mesh. (d) A novel view of textured mesh. Our proposed method can reconstruct the high-quality geometric mesh with a visually realistic texture.}
   \vspace{-4mm}
   \label{fig:onecol}
\end{figure}
Today's smartphones offer low-resolution depth maps \cite{arkitscenes} and valid poses. 
However, depth maps are very noisy and suffer from the limited range of depth sensors. Although this depth sensor can build a simple 3D structure such as a wall or floor, it cannot reconstruct objects with complex and varying shapes. 
Thus, the RGBD scanning methods \cite{izadi2011kinectfusion,niessner2013real,whelan2015elasticfusion,dai2017bundlefusion} are not suitable for these objects.
Instead of the depth sensor, multi-view stereo (MVS) algorithms \cite{furukawa2015multi,schonberger2016pixelwise,xu2020planar} reconstruct high-quality 3D geometry by matching feature correspondences across different RGB images and optimizing photometric consistency. While this pipeline is very robust in real-world environments, it misses the surface of low-textured areas \cite{xu2020planar}. Additionally, the resulting mesh generated by post-processing like Poisson reconstruction \cite{kazhdan2013screened} heavily depends on the quality of matching, and the accumulated errors in correspondence matching often cause severe artifacts and missing surfaces. Because of the cumulative error from the above pipeline, the texture mapping process \cite{waechter2014TexRecon,zhou2014colormapoptimization}
leads to undesirable results. Reconstructing high-fidelity texture and 3D geometry of real-world 3D objects remains an open challenge.

\noindent\textbf{Main Results:} 
In this paper, we present a practical method to capture a high-quality textured mesh of a 3D object in the wild, shown in Fig. \ref{fig:onecol}.
We first develop a smartphone app based on ARKit \cite{arkit} to collect images, LiDAR depths, and poses. 
Although the smartphone provides quite valid pose estimates, acquiring fine detail geometry and realistic texture requires highly accurate camera poses. Thus, we present an RGBD-aided Structure from Motion (SfM) which combines the advantages of both VIO \cite{flint2018visual} and SfM  \cite{schonberger2016structure}. Since ARKit based on VIO is robust to the degradation of visual information such as low-textured surface, we perform incremental triangulation with initial poses obtained from ARKit. We also propose a depth filtering method to handle noisy depth from the smartphone. These filtered depth points are exploited as an additional depth factor for bundle adjustment. Our RGBD-aided SfM can estimate highly precise camera poses due to the good initial poses and additional constraints from the filtered depth.

\begin{figure}[t]
  \centering
    \includegraphics[width=0.99\linewidth]{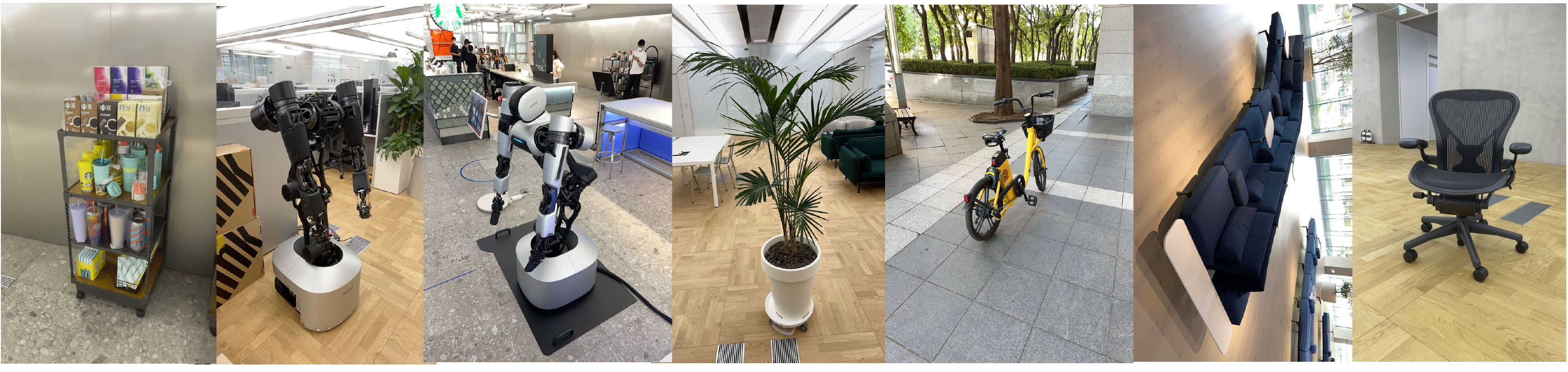}
    \vspace{-3mm}
   \caption{Example objects collected by a smartphone in the wild}
   \vspace{-4mm}
   \label{fig:every_objects}
\end{figure}

Our 3D geometry reconstruction process adopts a neural implicit representation \cite{wang2021neus,sun2022neural3dwild} for surface reconstruction with volumetric rendering. 
We observe that the neural implicit representation can perform more complete and smoother reconstruction than the classical methods which are known to be robust for 3D reconstruction in the wild.  
Furthermore, we introduce a new training method for neural implicit representations. 
In the early stage of training, we propose a regularization method that leverages the prior information from the classical MVS method. After obtaining the decent shape, we avoid using the prior information and generate sparse voxel octree to enable efficient sampling for training. Our training method can improve the performance of the neural implicit representations. Consequently, we show the generalization capabilities of our surface reconstruction method in real world objects collected by the smartphone.  

Given a mesh extracted from the trained neural implicit representation, we can run classical texture mapping algorithms \cite{zhou2014colormapoptimization,waechter2014TexRecon} to generate texture maps that often exhibit blurring artifacts and color misalignment. 
We propose applying differential rendering \cite{nvdiffrast} to fine-tune these texture maps obtained from classical texture mapping via a photometric loss. 
Compared to classical 3D reconstruction \cite{izadi2011kinectfusion,schonberger2016pixelwise,xu2020planar} and texture mapping \cite{waechter2014TexRecon,zhou2014colormapoptimization} approaches, our method shows a better ability to reconstruct the 3D model and produce realistic textures. We evaluate our approach on the data collected by our smartphone application.   
The main contributions of this paper are summarized as follows:
\begin{itemize}
    \item We present a unified framework to reconstruct the textured mesh using a smartphone.   
    \item We propose a depth filtering scheme for noisy depths and refine initial poses from ARKit by using bundle adjustment with depth factor.  
    \item Our pipeline builds on classical 3D reconstruction and texture mapping. We leverage a neural geometry representation to enable surface reconstruction of complex shapes and a differentiable rendering to generate high-fidelity texture maps.     
\end{itemize}


\section{Related Work}
\label{sec:relatedwork}
\noindent\textbf{Classical 3D Reconstruction}
Traditional Structure from Motion (SfM) systems \cite{wu2013towards,schonberger2016structure} have been popular as a vision-based 3D reconstruction due to their robustness to various scenarios \cite{snavely2008modeling,snavely2006photo,widya2020stomach}. The SfM systems extract and match features \cite{lowe2004distinctive} of adjacent images of the same scene and generate the camera poses and sparse 3D points via triangulation \cite{hartley2003multiple}. Then, Multi-view Stereo (MVS) algorithms \cite{vogiatzis2005multi,furukawa2015multi} compute the dense depth maps of each scene with multiple calibrated images. 
PatchMatch-based MVS \cite{gipuma,schonberger2016pixelwise,xu2020planar,xu2022multi} methods are popular because they are proper for depth and normal optimization. 
After that, surface reconstruction methods \cite{kazhdan2006poisson,labatut2009robust,kazhdan2013screened} can be applied to reconstruct meshes from point clouds. However, all these methods often fail to reconstruct the 3D geometry of low-textured areas. RGB-D Reconstruction techniques fuse many overlapping depth maps into a single 3D model. Starting from the classical TSDF-fusion \cite{curless1996volumetric}, several methods \cite{izadi2011kinectfusion,niessner2013real,whelan2015elasticfusion,dai2017bundlefusion} produce voxel-based TSDF with commodity-level depth cameras such as Kinect. 
However, these methods have difficulty handling outliers and thin geometry due to noisy depth measurements. Due to the noise and limited range of the LiDAR sensor in smartphone, RGB-D reconstruction methods fail to estimate the camera odometry.   

\noindent\textbf{3D Reconstruction with differentiable rendering}
Recently, NeRF \cite{mildenhall2021nerf} and its variants \cite{martin2021nerfinthewild,yu2021plenoctrees,muller2022instant,zhang2020nerf++} have used volumetric implicit representations for rendering and learn \(\alpha\)-compositing of a radiance field along rays. These methods yield remarkable quality of novel view synthesis results and do not need ground truth masks. However, 3D geometry from volumetric representations is far from satisfactory due to their volumetric properties \cite{zhang2020nerf++}. Some methods \cite{yariv2020IDR,mvsdf,niemeyer2020differentiable} reconstruct  the surface by encoding the network as an implicit function such as a signed distance function (SDF) or occupancy function. Many hybrid approaches \cite{wang2021neus,oechsle2021unisurf,yariv2021volsdf,azinovic2022neuralrgbdsurfacerecon,NeuralWarp} unify surface and volumetric representations by regarding surface as defining volumes near the surface, enabling accurate surface reconstruction without masks. Their follow-up works \cite{mvsdf,sparseneus,neuris,manhattansdf} import prior information from the learning-based network. We avoid exploiting prior hints obtained from learning-based models because they are expensive to acquire new training datasets and time-consuming to train other neural networks.   

\noindent\textbf{Classical Texture Optimization}
The matching between texture and geometry is a significant step for building realistic 3D models. Lempitsky et al. \cite{lempitsky2007seamless} apply the pairwise Markov Random Field to choose an optimal view for each face as texture. Waechter et al. \cite{waechter2014TexRecon} present a global color adjustment to mitigate the visual seams between face textures. Zhou et al. \cite{zhou2014colormapoptimization} optimize the camera poses using a color consistency measurement and geometric error via local image warping to obtain sharp texture. Bi et al. \cite{bi2017patch} employ a patch-based image synthesis to produce a texture per each face to optimize the camera pose and geometric error.       

\noindent\textbf{Texture Learning}
Recently, learning-based methods \cite{Kanazawa_2018_ECCV,Kanazawa_2018ECCV2,huang2020adversarial,10.1145/3306346.3323035} have been proposed to generate textures for a 3D model.
Huang et al. \cite{huang2020adversarial} introduce adversarial learning for texture generation in RGB-D scans. 
Thies et al. \cite{10.1145/3306346.3323035} store neural textures for a known mesh with given UV mapping and decode them to color for novel views. Some works \cite{NeuTex,neumesh} learn neural texture with neural volumetric rendering. Differentiable rendering methods \cite{henderson2020leveraging,9915626,munkberg2021nvdiffrec,Goel_2022_CVPR} usually focus on estimating surface
radiometric properties from images. These methods can learn to predict not only texture but also geometry via image loss with ground truth masks. Unfortunately, differentiable rendering methods heavily rely on pixel-level accurate masks. Acquiring accurate ground-truth masks in the real-world environment is a difficult task \cite{Kirillov_2020_CVPR}. Therefore, we apply the differentiable rendering to fine-tune texture maps with a fixed mesh.

\section{Our Method}
\label{sec:ourmethod}
Given images, low-resolution depths and corresponding camera poses from ARKit, our goal is to reconstruct both the texture and geometry of real-world 3D objects. 
In Fig. \ref{fig:framework}, our pipeline consists of three stages: RGBD-aided Structure from Motion (SfM), Geometry Reconstruction, and Texture Optimization.  
Since both depth maps and camera poses provided by ARKit are very noisy, we construct a SfM system to filter out depth maps and refine poses. We then formalize our task as training a neural model for zero-level set of a signed distance function (SDF) guided by classical 3D reconstruction. After training, we extract the mesh with Marching Cubes \cite{marchingcube} and apply classical texture reconstruction \cite{waechter2014TexRecon} to generate initial texture maps. Next, we exploit differentiable rendering \cite{nvdiffrast} to finetune the texture maps with the supervision of multi-view images.      

\subsection{RGBD-aided Structure from Motion} 
\label{sec:Tracking}
The goal of this section is to refine initial poses from ARKit by performing a bundle adjustment with depth maps from the smartphone. Unlike simple structures like walls or floors, 3D reconstruction for capturing a high level of detail is required to refine inaccurate poses from ARKit. 

\noindent\textbf{Preprocessing}
We introduce two processes to obtain reliable depth estimates: single-view filtering and multi-view filtering. Single-view filtering utilizes a confidence map provided by ARKit, which has the same resolution as the depth map. Each pixel in the confidence map has 0, 1, or 2 with the value 0 as least confident and 2 serving as most confident with respect to depth-value accuracy. We choose depth pixels corresponding to 1 or 2 in the confidence map. A main concept of the multi-view filtering is to remove outliers in terms of the reprojection error. 
Given the reference image $I_{ref}$ with depth $D_{ref}$, the source image $I_{src}$ with depth $D_{src}$, the depth map $D'_{src}$ can be transformed from the source view to the reference view with accurate pixel-level correspondence. Then, the depth of pixel $p$ in the reference image will be considered as an inlier if the difference $\delta_{depth}$ between $D_{src}$ and $D'_{src}$ is smaller than a small constant value $\epsilon$. Then, assuming a set of neighboring N source images, we compute the number of inliers which fulfils $\delta_{depth} < \epsilon$ for each source image, and keep depth pixels if the number of inliers is greater than a certain threshold.        
Finally, we apply both single-view and multi-view filtering to remove all depth outliers. 

\begin{figure}[t]
    \centering
    \includegraphics[width=\linewidth]{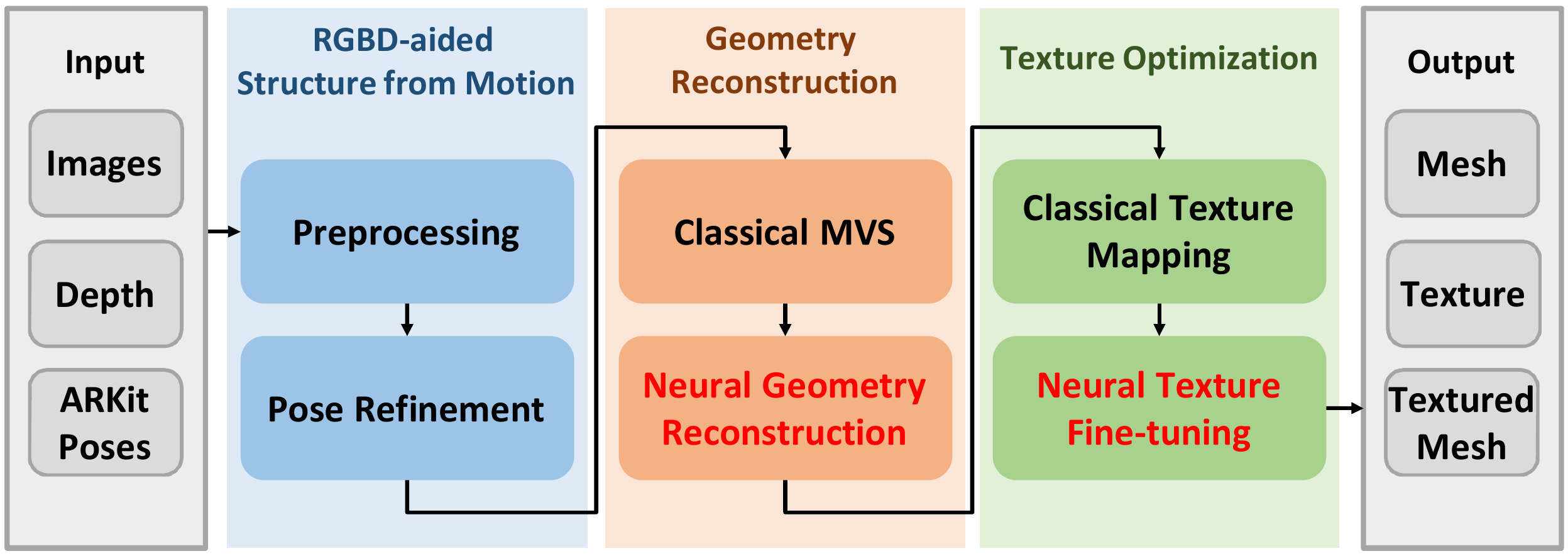}
    \vspace{-6mm}
    \caption{
    Overview of our pipeline. It is comprised of three steps: RGBD-aided Structure from Motion (Sec. \ref{sec:Tracking}), Geometry Reconstruction (Sec. \ref{sec:Geometry}), and Texture Optimization (Sec. \ref{sec:Texture}). Our method builds on classical 3D reconstruction and texture mapping. To overcome the limitations of classical methods, we apply a neural geometry representation and neural texture fine-tuning to update 3D geometry and texture maps.       
    }
    \label{fig:framework}
    \vspace{-4mm}
\end{figure}

\noindent\textbf{Pose Refinement}
We consider camera poses obtained from ARKit as initial poses and exploit matching correspondences to perform incremental triangulation \cite{schonberger2016structure}. After combining filtered depth points from LiDAR and 3D points from incremental triangulation, we propose a depth factor and run bundle adjustment with the depth factor and reprojection error. Let $P_{i}$ denote the camera parameter of the $i$-th camera and $X_{j}$ denote point parameters seen by the $i$-th camera.  We minimize the proposed joint objective function $E$, which is formulated as follows:   
\begin{equation} \label{eq2}
    E = \sum_{i}\sum_{j}\rho_{i}(\parallel \phi(P_{i}, X_{j}) - x_{ij} \parallel^2) + \frac{1}{\eta}|D_{i} -  z_{i}|,
\end{equation}
where $\Phi$ is the function that projects points to the image plane, and $\rho$ denotes the Cauchy function as the robust loss function. $x_{ij}$ is the feature point corresponding to the 3D point $X_{j}$ seen by the $i$-th camera. $z_{i}$ is the depth value of points projected from the 3D point $X_{j}$ and $D_{i}$ is the filtered depth of pixels corresponding to $z_{i}$. We define $\eta$ as sensor measurement noise and use it to down-weight noisy depth. We notice that the depth factor has two benefits. First, additional information from the depth sensor is a useful constraint for accurate pose estimation and reliable triangulation. Furthermore, this depth factor guides the scale of the SfM to follow the metric scale of the depth sensor.

\subsection{Geometry Reconstruction}
\label{sec:Geometry}
\noindent\textbf{Neural Geometry Representations} 
Instead of classical 3D reconstruction, we adopt NeuS \cite{wang2021neus}, which combines the advantages of both volume and surface rendering. In other words, it can extract accurate surfaces even without pixel-level masks. Our scanned object is represented by two Multi-layer Perceptrons (MLPs): geometry network $f_{sdf}$, which maps a 3D point $x \in \mathbb{R}^3$ to a signed distance, and color network $f_{color}$ which converts a 3D point $x$ and a viewing direction $v \in \mathbb{R}^3$ to a color $c$.  The 3D geometry of an object is represented implicitly as an SDF with zero-level set \{$x \in \mathbb{R}^3 | f_{sdf}(x)=0)$\}. These two parameters of geometry and color are learned by the volume rendering technique. For a ray \{$x(t)=o + tv | t \geq 0$\} with camera center $o$ and viewing direction $d$, the color is rendered with $n$ sampled points along the ray as follows:
\begin{equation} \label{eq:neuralsurfacerecon}
    \hat{C} = \sum_{i=1}^{n}T_{i}\alpha_{i}c(x(t), v),
\end{equation}
where $T_{i} =\prod_{j=1}^{i-1} (1-\alpha_{j})$ is the accumulated trasnmittance, and $\alpha_{i}$ is the discrete opacity. Here, $\alpha_{i} = 1 - \text{exp}(-\int_{t_{i}}^{t_{i+1}} \rho(t) \,dt)$, and $\rho(t)$ is opaque density function borrowed from the definition in NeuS \cite{wang2021neus}.  

\noindent\textbf{Training Process}
RGBD-aided SfM already estimates both poses and sparse depths by matching feature correspondences. Given this prior information, we leverage an MVS algorithm \cite{xu2020planar} to estimate the depth and normal for all images with less cost. It is worth noting that we avoid using learning-based MVS algorithms because they are not robust to diverse real-world scenes and are hard to evaluate reliability.  In this section, we modify the training process of NeuS to avoid losing fine details of complex shape. Our training process is divided into two stages. In the first stage, we train a neural radiance field with a prior depth and normal obtained from the MVS \cite{xu2020planar} algorithm to supervise the training process. Specifically, we sample $K$ pixels $p_{k}$ and their corresponding color values $I(p_{k})$ in each iteration. Here, we propose a regularization method that relies on prior depth $D$ and normal $N$ from MVS algorithm. We unproject depth under the given pose to obtain 3D point $X(p_{k})$. Then, we use this 3D point and its corresponding normal $N(p_{k})$ from MVS to regularize the geometry network by minimizing     
\begin{equation} \label{eq:regularization}
    \begin{split}
    L_{reg} = \sum_{k}\:\:&w_{d}|f_{sdf}(X(p_{k}))| \:\:\:+ \\
    & w_{n}(1 - \langle\nabla_{p} f_{sdf}(X(p_{k})), N(p_{k})\rangle),
    \end{split}
\end{equation}
\begin{figure}[t]
    \centering
    \includegraphics[width=\linewidth]{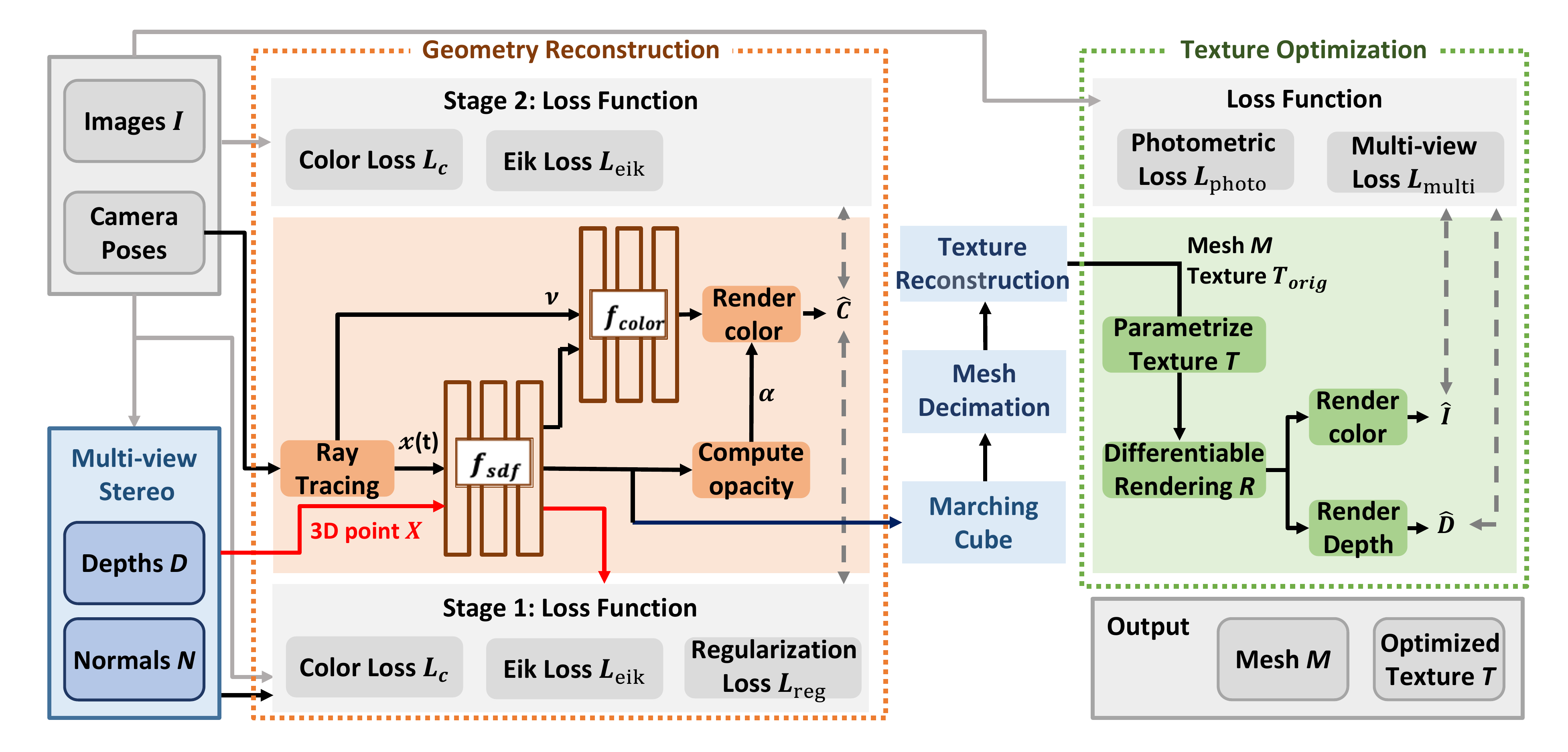}
    \vspace{-7mm}
    \caption{
    Overview of Geometry Reconstruction (Section \ref{sec:Geometry}) and Texture Optimization (Section \ref{sec:Texture}). Our goal in geometry reconstruction is to train geometry network $f_{sdf}$ and color network $f_{color}$ via volumetric rendering with two training steps. In step 1, we leverage prior geometric information from the MVS algorithm \cite{xu2020planar}. Then, our method utilizes a sparse voxel octree to efficiently sample points near the surface. In texture optimization, we first utilize a classical texture reconstruction to generate texture maps for a given mesh. Then, our method applies a differentiable rendering to fine-tune texture maps. }
    \label{fig:framework}
    \vspace{-4mm}
\end{figure}
where $w_{d}$ and $w_{n}$ are weights for each term. This regularization will guide the SDF to learn explicit supervision from the MVS algorithm. Although this regularization is helpful for learning prior geometry, it is susceptible to the noisy results derived from the MVS algorithm which tends to lose fine details and texture-less regions. Thus, we will only use this regularization in the first stage.  
The color of each pixel can be computed by E.q. \ref{eq:neuralsurfacerecon}. The color loss $L_{c}$ is defined as   
\begin{equation} \label{eq:neuralsurfacerecon}
    L_{c} = \sum_{k}|\hat{C}(p_{k}) - I(p_{k})|.
\end{equation}
The Eikonal loss \cite{eik} to regularize the SDF is formulated as 
\begin{equation} \label{eq:eikonal}
    L_{eik} = \sum_{k,i}(\parallel\nabla f_{sdf}(p_{k,i})\parallel_{2} - 1)^2.
\end{equation}
The overall loss function $L$ in the first stage is defined as 
\begin{equation} \label{eq:neuralsurfacerecontotal}
    L = L_{c} + L_{eik} + L_{reg}.
\end{equation}
In the second stage, inspired by \cite{sun2022neural3dwild}, we adopt a sparse voxel octree to guide the sampling process and capture fine details. 
After the first stage, we can extract a triangular mesh from the SDF via the Marching Cube\cite{marchingcube} and define a sparse voxel volume based on this mesh. Then, based on the sparse voxel volume, we can avoid redundant point samples outside the sparse voxel and focus on point samples near the surface samples. Here, our total loss is $L = L_{c} + L_{eik}$, which is the same as E.q. \ref{eq:neuralsurfacerecontotal} except for $L_{reg}$. To capture fine details, we will not use $L_{reg}$ in E.q. \ref{eq:neuralsurfacerecontotal}. 

\noindent\textbf{Mesh Simplification}
A highly detailed 3D mesh from the SDF network with marching cube algorithms \cite{marchingcube} leads to enormous memory requirements. To obtain a lightweight mesh, we explore popular quadric mesh simplification \cite{garland1997surface}. Although the mesh simplification generated small holes, the lightweight mesh enables us to reduce the computation complexity of the texture mapping process. We set the decimation ratio to 0.4\%.    

\subsection{Texture Optimization}
\label{sec:Texture}
\noindent\textbf{Classical Texture Reconstruction} For a triangular mesh $M$ provided by the previous section, we first utilize Waechter et al. \cite{waechter2014TexRecon} with poses and images across different views to construct a texture map $T_{orig}$. Compared to the color network from Section \ref{sec:Geometry}, this classical method \cite{waechter2014TexRecon} can acquire visually realistic texture images. However, it cannot avoid texture bleeding on the boundary of different views and blurring artifacts.   

\noindent\textbf{Texture Fine-tuning}
To tackle blurring and texture bleeding artifacts, we propose a fine-tuning step by optimizing the texture map. Here, we parameterize a texture map $T$ obtained from classical texture reconstruction. Given a 3D Mesh $M$ with the 2D texture map $T$ and multi-view images $I_{i}$ with the corresponding camera poses $P_{i}$, we leverage a differentiable renderer $R$ to generate an image $\hat{I}_{i}$ and depth $\hat{D}_{i}$, \ie $\hat{I}_{i}, \hat{D}_{i} = R(M, P_{i})$. A differentiable renderer $R$ \cite{nvdiffrast} performs mesh rasterization and then texture sampling to render an image $\hat{I}_{i}$. In other words, we project texture onto the image plane and compare it to the corresponding view of the image. Furthermore, we compare the rendered image with a different view image by minimizing multi-view photometric loss.   
We use the combination of the L1 and SSIM \cite{wang2004image} as the photometric loss $L_{photo}$ in image space between the rendered image $\hat{I}_{i}$ and the reference image $I_{i}$ as follows:
\begin{equation} \label{eq:textureupdate1}
    L_{photo} = \sum_{i} (1-\alpha)\parallel\hat{I}_{i} - I_{i}\parallel + \frac{\alpha}{2}(1 - SSIM(\hat{I}_{i}, I_{i})),
\end{equation}
where $\alpha$ is set to 0.85. 
Additionally, the differentiable render estimates depth $\hat{D}_{i}$ by interpolating $z$ coordinates of each vertex. With the estimated depth $\hat{D}_{i}$ and intrinsic parameter $K$, we can generate a synthesized frame $I'_{i}$ by reprojecting adjacent frames $I_{i'}$ to the current frame $I_{i}$ with the camera intrinsic matrix $K$, the estimated depth $\hat{D}_{i}$, and the relative pose $P_{i\rightarrow i'}$. The multi-view photometric loss $L_{multi}$ similar to Eq. \ref{eq:textureupdate1} is formulated as: 
\begin{equation} \label{eq:textureupdate2}
\begin{split}
    I_{i}^{\prime}(p)  &= I_{i^{\prime}}\langle\pi (K P_{i \rightarrow i^{\prime}} \hat{D}_{i}(p) K^{-1} \tilde{p})\rangle, \\
    L_{multi} = &(1 - \alpha) \parallel I_{i}^\prime - I_{i}\parallel + \frac{\alpha}{2}(1 - SSIM(I^{\prime}_{i}, I_{i})),
\end{split}
\end{equation}
where  $\tilde p$ is the homogeneous coordinate of $p$, $\pi$ means projection from homogeneous to image coordinates, and $\langle \cdot \rangle$ indicates the bilinear sampling function. 
We define our total loss $L$ to be 
\begin{equation} \label{eq:textureupdate3}
L = L_{photo} + L_{multi}.
\end{equation}
Finally, the differentiable rendering enables the loss gradients to back-propagate to update the textures parameters. Since our method does not require ground truth masks, we avoid considering optimization for the geometry. We observe that joint optimization with geometry and texture fails to converge without the ground truth mask.  

\section{Experiments}
\subsection{Experiment Settings}
\noindent\textbf{Data Collection:}
Our smartphone application, running on an iPhone 13 Pro using ARKit 6, can record various sensor outputs: synchronized images, poses, confidence maps, dense depth maps from the LiDAR scanner, and IMU sensor information. We built two types of datasets: ARKit-video and AR-capture. ARKit-video collects long video sequences similar to datasets constructed from RGBD-SLAM \cite{zollhofer2018state}. The second type is collected by our smartphone application based on Object Capture \footnote{https://developer.apple.com/videos/play/wwdc2021/10076/}. AR-capture is a multi-view dataset of objects consisting of high-quality 4K images with depth and gravity data. The ARKit-video collects thousands of frames sampled from the video with resolution of 1280x720. In contrast, AR-capture mostly contains less than fifty images at 4032x3096 pixels. We collect 11 objects: \textit{robot arm 1}, \textit{robot arm 2}, \textit{plant}, \textit{tree}, \textit{sofa}, \textit{bike}, \textit{recliner chair}, \textit{cafe stand}, \textit{delivery robot}, \textit{office chair}, and \textit{camera stand}.  
We randomly select 70 \% as the training images and utilize the 30 \% as test images for evaluating the novel view synthesis. 

\noindent\textbf{DTU Dataset:}
The DTU dataset \cite{DTUdataset} is a MVS dataset that provides reference point clouds acquired with laser sensor in the controlled lab. Since AR-capture and ARKit-video do not include 3D point clouds acquired with specialized hardware, we use the DTU dataset to evaluate the performance of geometry reconstruction. We follow the evaluation of NeuS \cite{wang2021neus} and exploit the chamfer distance as metric.  

\noindent\textbf{Implementation Details:}
We build the RGBD-aided Structure from Motion based on COLMAP \cite{schonberger2016structure}, an existing SfM tool. Instead of SIFT \cite{lowe2004distinctive}, our system uses learning-based features \cite{detone18superpoint,revaud2019r2d2,Sarlin_2020_CVPR} for feature extraction and matching to avoid failure in a featureless scene. ARKit offers an interface for accessing LiDAR data as a depth image. 
Following the work \cite{wang2021neus}, the SDF network $f_{sdf}$ is implemented as an 8-layer MLP a hidden dimension of 256, and geometric initialization \cite{atzmon2020sal} for the network parameters. We construct the color network $f_{color}$ as MLPs with 4-layer MLP with hidden dimension of 256. We optimize networks in geometry reconstruction with Adam \cite{kingma2014adam} and an initial learning rate of 5e-4. We train the first step for 60K iterations and the second step for 140K iterations. In terms of objects with complex shape, it takes 240K iterations for second step. In Texture Optimization, we utilize the public github code for classical method \cite{waechter2014TexRecon}. Then, for texture fine-tuning, the Adam optimizer is used with the learning rate for the texture parameter. This step takes 3000 epochs to fine-tune the texture parameter.

\begin{figure}[t]
    \vspace{-2mm}
    \centering
    \includegraphics[width=0.99\linewidth]{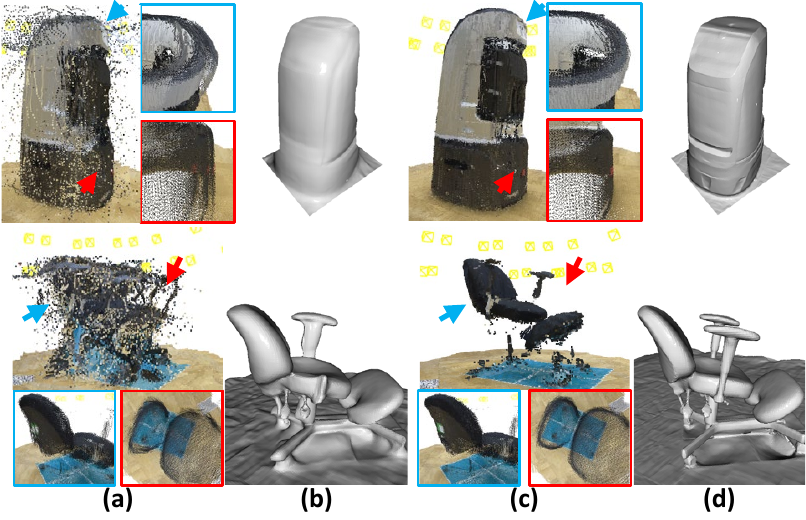}
    \vspace{-2mm}
    \caption{We show the benefits of RGBD-aided Structure from Motion. (a) illustrates the 3D point clouds obtained from initial depths with initial poses. In (c), we accumulate the 3D point clouds (by using filtered depths and refined poses) from RGBD-aided SfM. For a detailed description, we visualize the zoomed-in blue and red regions using filtered depth. Initial depth maps and poses from ARkit are input to the geometry reconstruction in Section \ref{sec:Geometry} and its results are shown in (b). The results of our pipeline in (d) lead to higher quality reconstructions than (b) which does not apply RGBD-aided SfM.    
    }
    \label{fig:only_arkit}
    \vspace{-4mm}
\end{figure}

\subsection{Experimental Results}
In Fig. \ref{fig:only_arkit}, we visualize the effectiveness of RGBD-aided Structure from Motion in Section \ref{sec:Tracking} on our dataset. Figure \ref{fig:only_arkit} (a) represents the 3D point clouds by initial depth maps and poses obtained from ARKit. Unfortunately, initial sensor data is too noisy to perform an accurate 3D reconstruction. After multi-view depth filtering as preprocessing, zoomed-in regions in (a) show the point clouds by reprojecting filtered depth maps with initial poses. Compared to initial data, multi-view depth filtering can handle noisy depth measurements. Figure \ref{fig:only_arkit} (c) provides the results of preprocessing and pose refinement. Pose refinement enables us to globally align 3D reconstruction without noticeable camera drift and with better local quality. For qualitative comparison, we take as input initial sensor data from ARKit and perform geometry reconstruction in Section \ref{sec:Geometry} in Fig. \ref{fig:only_arkit} (b). Compared to our experimental results in Fig. \ref{fig:only_arkit} (d), it is difficult to achieve comparable quality and better reconstruction. Consequently, our proposed RGBD-aided SfM significantly improves geometry reconstruction performance.

\begin{figure}[t]
    \vspace{-2mm}
    \centering
    \includegraphics[width=0.85\linewidth]{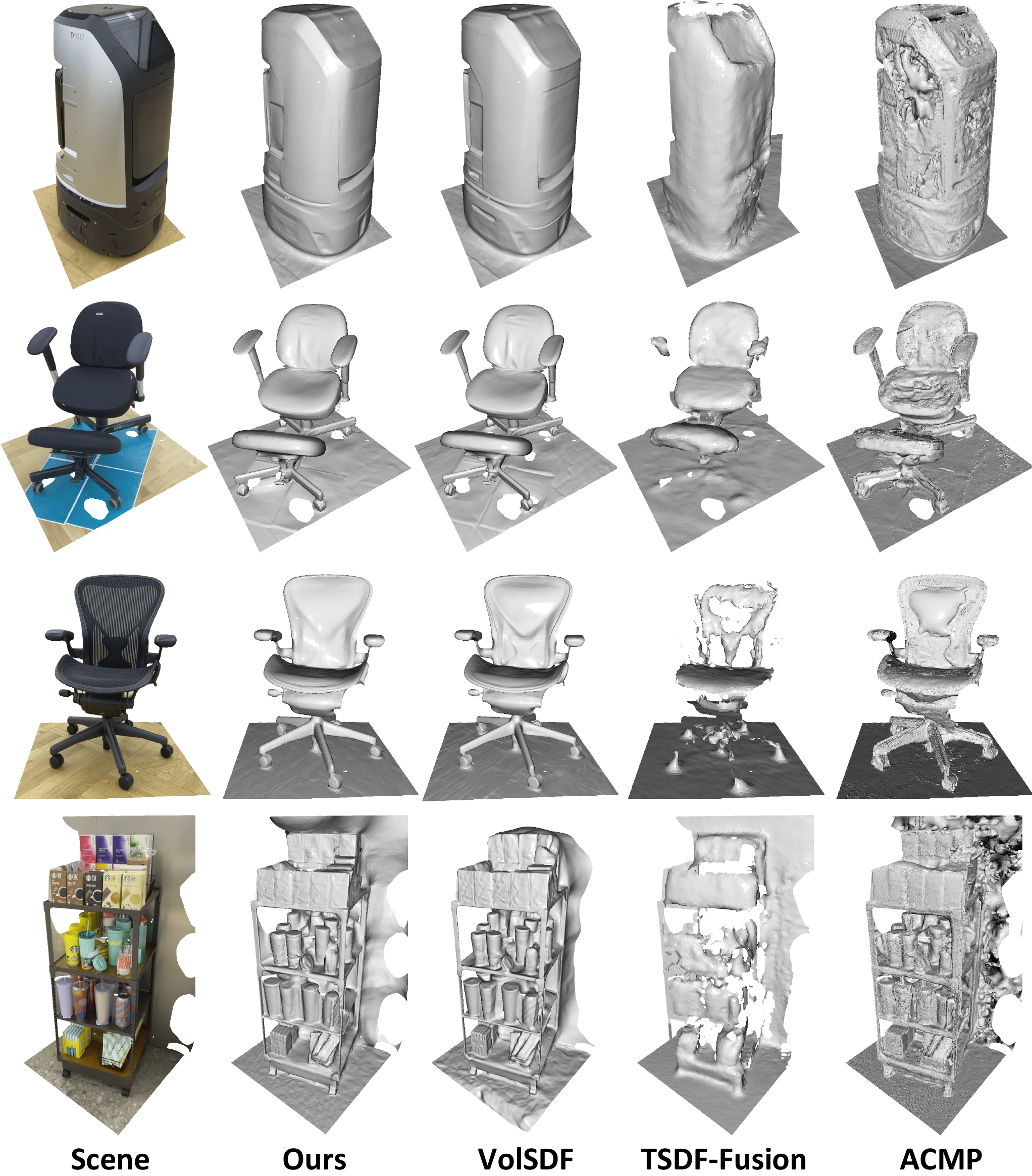}
    \vspace{-2mm}
    \caption{Reconstruction of our method, VolSDF \cite{yariv2021volsdf}, TSDF-Fusion \cite{izadi2011kinectfusion}, and a mesh from point clouds of ACMP \cite{xu2020planar} processed with screened Poisson surface reconstruction \cite{kazhdan2013screened}. Note that the neural surface reconstruction is more perceptually convincing in the ARKit-video dataset.}
    \label{fig:arkit_geometry_comparison}
    \vspace{-4mm}
\end{figure}

Figure \ref{fig:arkit_geometry_comparison} qualitatively compares our method to VolSDF \cite{yariv2021volsdf}, TSDF-fusion \cite{izadi2011kinectfusion}, and ACMP \cite{xu2020planar}. We apply screened Poisson surface reconstruction \cite{kazhdan2013screened} to generate a mesh from point clouds of ACMP. Among all four algorithms, we apply the same pose estimation algorithm from Section \ref{sec:Tracking} for fair comparison. TSDF-fusion, which requires sufficient scene overlap, exploits more frames than our method and ACMP. Due to the low-textured regions, the surface of ACMP is very noisy. TSDF-fusion utilizes depth maps obtained from low-resolution LiDAR. Here, since this low-resolution depth sensor is very noisy, we only use filtered depth maps after preprocessing in Section \ref{sec:Tracking}. However, this model still cannot reconstruct either thin or challenging structures accurately (e.g., a chair armrest).  Both implicit neural surface reconstruction approaches show comparable performance, and both are superior to classical 3D reconstruction. We observe that the difference between neural geometric representations (Ours and VolSDF) is negligible unlike pose quality. It is worth noting that using accurate poses as input significantly improves the 3D reconstruction quality as shown in Fig. \ref{fig:only_arkit}. 

\begin{figure*}[t]
    \vspace{-2mm}
    \centering
    \includegraphics[width=0.73\linewidth]{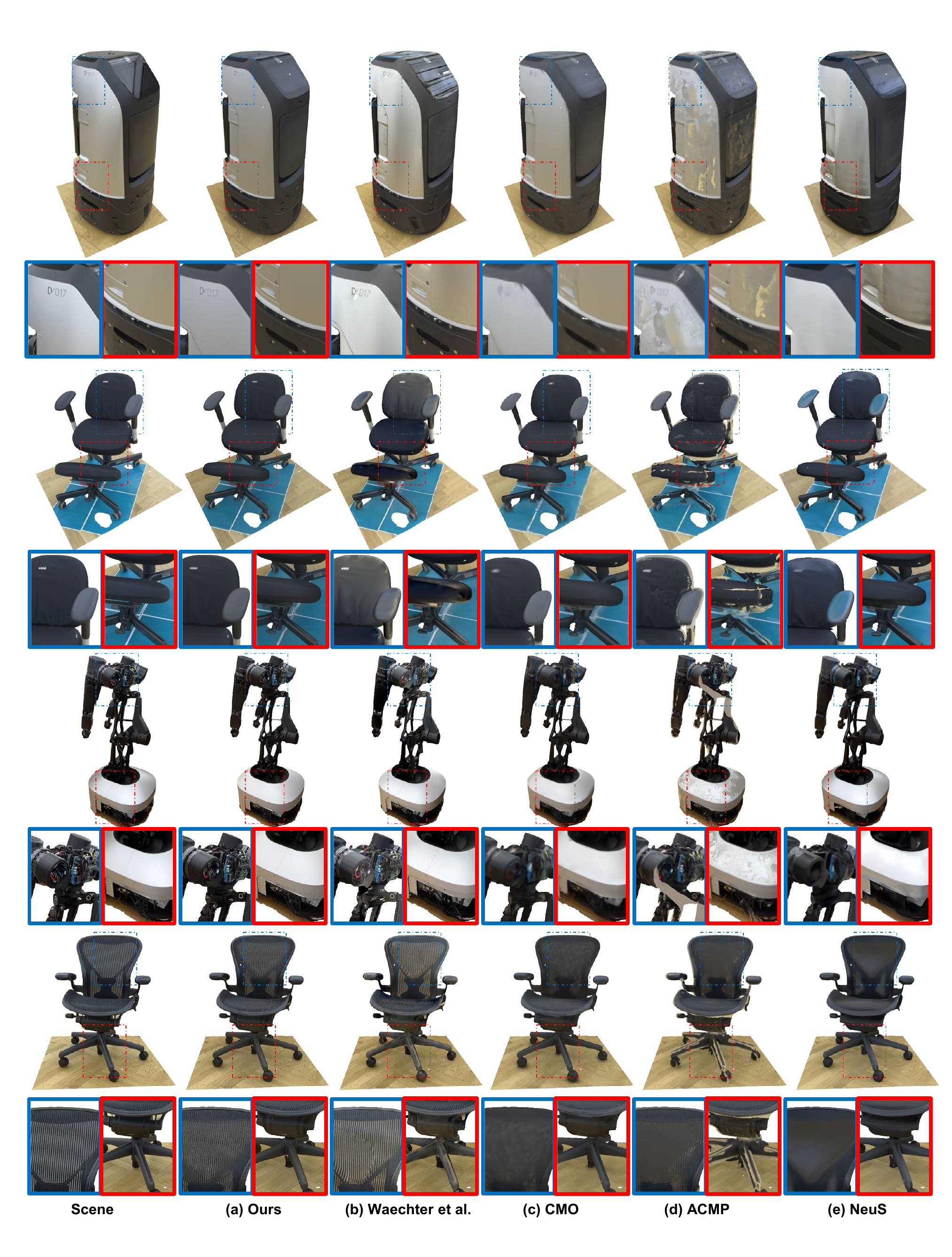}
    \vspace{-6mm}
    \caption{Qualitative comparison between (a) Ours, (b) Waechter et al. \cite{waechter2014TexRecon}, (c) CMO \cite{zhou2014colormapoptimization}, (d) ACMP \cite{xu2020planar}, and (e) NeuS \cite{wang2021neus}. Compared to textured mesh (3-6 columns) obtained from different algorithms, the result of our method is perceptually closer to the original scene.  Additional qualitative results are added in the supplementary material due to the limited space.}
    \label{fig:arkit_texture_comparison}
    \vspace{-4mm}
\end{figure*}

\begin{table*}[t]
    \centering
    \resizebox{0.8\textwidth}{!}{\normalsize
    \begin{tabular}{l|c|ccc|ccc|ccc|ccc}
    \toprule
    \multirow{2}{*}{Category}&\multirow{2}{*}{Data}&\multicolumn{3}{c|}{NeuS \cite{wang2021neus}} &\multicolumn{3}{c|}{CMO \cite{zhou2014colormapoptimization}} & \multicolumn{3}{c|}{Waechter et al. \cite{waechter2014TexRecon}} & \multicolumn{3}{c}{Ours} \\
    & & PSNR \(\uparrow\) & SSIM \(\uparrow\) & LPIPS \(\downarrow\) & PSNR \(\uparrow\) & SSIM \(\uparrow\) & LPIPS \(\downarrow\)  & PSNR \(\uparrow\) & SSIM \(\uparrow\) & LPIPS \(\downarrow\) & PSNR \(\uparrow\) & SSIM \(\uparrow\) & LPIPS \(\downarrow\)\\
    \midrule
    robot arm 1 & Ac & 20.292 & 0.889 & 0.083 & 18.473 & 0.891 & 0.083 & 19.909 & 0.872 & 0.060 & 20.049 & 0.871 & 0.057 \\
    robot arm 2 & Ac &  20.741 & 0.882 & 0.086 & 21.107 & 0.883 & 0.092 & 20.190 & 0.862 & 0.067 & 20.652 & 0.858 & 0.062 \\
    plant & Ac & 16.748 & 0.642 & 0.295 & 18.462 & 0.627 & 0.316 & 17.921 & 0.635 & 0.186 & 19.970 & 0.691 & 0.146 \\
    tree & Ac & 18.409 & 0.869 & 0.117 & 18.448 & 0.870 & 0.127 & 17.518 & 0.858 & 0.096 & 18.331 & 0.858 & 0.102 \\
    sofa & Ac & 19.910 & 0.872 & 0.139 & 21.170 & 0.840 & 0.136 & 19.759 & 0.829 & 0.113 & 21.608 & 0.894 & 0.109 \\
    bike & Ac & 20.971 & 0.778 & 0.183 & 21.311 & 0.778 & 0.206 & 19.822 & 0.740 & 0.088 & 19.943 & 0.737 & 0.078 \\
    \midrule
    Mean & Ac & 19.512 & 0.822 & 0.151 & 19.825 & 0.815 & 0.160 & 19.186 & 0.799 & 0.101 & \textbf{20.095} & \textbf{0.818} & \textbf{0.092} \\
    \midrule
    \midrule
    recliner chair & Av & 24.920 & 0.893 & 0.099 & 24.761 & 0.876 & 0.112 & 23.692 & 0.864 & 0.054 & 25.045 & 0.896 & 0.056 \\
    cafe stand & Av & 20.498 & 0.779 & 0.173 & 22.934 & 0.828 & 0.154 & 17.915 & 0.798 & 0.112 & 22.634 & 0.818 & 0.095  \\
    delivery robot & Av & 22.412 & 0.897 & 0.126 & 23.549 & 0.908 & 0.128 & 25.246 & 0.929 & 0.073 & 23.451 & 0.906 & 0.086 \\
    office chair & Av & 20.722 & 0.892 & 0.096 & 21.120 & 0.899 & 0.103 & 20.315 & 0.865 & 0.094 & 20.982 & 0.892 & 0.083 \\
    camera stand & Av & 21.901 & 0.914 & 0.069 & 21.886 & 0.907 & 0.074 & 21.121 & 0.899 & 0.059 & 22.208 & 0.919 & 0.060 \\
    \midrule
    Mean & Av & 22.092 & 0.875 & 0.113 & 22.850 & 0.884 & 0.115 & 21.658 & 0.871 & 0.079 & \textbf{22.868} & \textbf{0.886} & \textbf{0.076} \\
    \bottomrule
    \end{tabular}
    }
    \vspace{-2mm}
    \caption{Quantitative comparison on our dataset. Ac and Av denote the AR-catpure dataset and ARKit-video dataset, respectively. We improve the LPIPS metric by 8.9\% in AR-capture and by 3.8\% in ARKit-video. 
    }
    \vspace{-2mm}
    \label{table: metric}
\end{table*}

\begin{table*}[t]
    \centering
    \resizebox{0.8\textwidth}{!}{\normalsize
    \begin{tabular}{l|c|c|c|c|c|c|c|c|c|c|c|c|c|c|c|c}
    \toprule
    Method & scan24 & scan37 & scan40 & scan55 & scan63 & scan65 & scan69 & scan83 & scan97 & scan105 & scan106 & scan110 & scan114 & scan118 & scan122 & Mean \\ 
    \midrule
    COLMAP \cite{schonberger2016pixelwise} with trim=7 & 0.45 & 0.91 & 0.37 & 0.37 & 0.90 & 1.00 & 0.54 & 1.22 & 1.08 & 0.64 & 0.48 & 0.59 & 0.32 & 0.43 & 0.45 & 0.65 \\
    \midrule
    UNISURF \cite{oechsle2021unisurf} & 1.32 & 1.36 & 1.72 & 0.44 & 1.35 & 0.79 & 0.80 & 1.49 & 1.37 & 0.89 & 0.59 & 1.47 & 0.46 & 0.59 & 0.62 & 1.02 \\ 
    NeuS \cite{wang2021neus} & 1.37 & \textbf{1.21} & 0.73 & \textbf{0.40} & 1.20 & 0.70 & 0.72 & 1.01 & 1.16 & 0.82 & 0.66 & 1.69 & \textbf{0.39} & 0.61 & 0.51 & 0.87 \\ 
    VolSDF \cite{yariv2021volsdf} & 1.14 & 1.26 & 0.81 & 0.49 & 1.25 & 0.70 & 0.72 & 1.29 & 1.18 & \textbf{0.70} & 0.66 & 1.08 & 0.42 & 0.61 & 0.55 & 0.86 \\
    \midrule
    Ours without 2nd step & 1.04 & 1.61 & 0.67 & 0.64 & 0.88 & 0.70 & 0.62 & 1.14 & 1.10 & 0.89 & 0.66 & \textbf{1.02} & 0.50 & 0.58 & 0.51 & 0.84\\ 

    Ours & \textbf{0.96} & 1.37 & \textbf{0.67} & 0.60 & \textbf{0.88} & \textbf{0.70} & \textbf{0.45} & \textbf{1.14} & \textbf{1.10} & 0.84 & \textbf{0.66} & 1.11 & 0.50 & \textbf{0.58} & \textbf{0.50} & \textbf{0.80}\\ 
    \bottomrule
    \end{tabular}
    }
    \vspace{-2mm}
    \caption{ Quantitative comparison on DTU \cite{DTUdataset}. Chamfer distance is employed as the evaluation metric. COLMAP is superior to other neural geometry reconstructions. Our training process can improve the NeuS by 8 \%.    
    }
    \vspace{-3mm}
    \label{table:dtu_metric}
\end{table*}

\begin{figure}[t]
    \centering
    \includegraphics[width=0.85\linewidth]{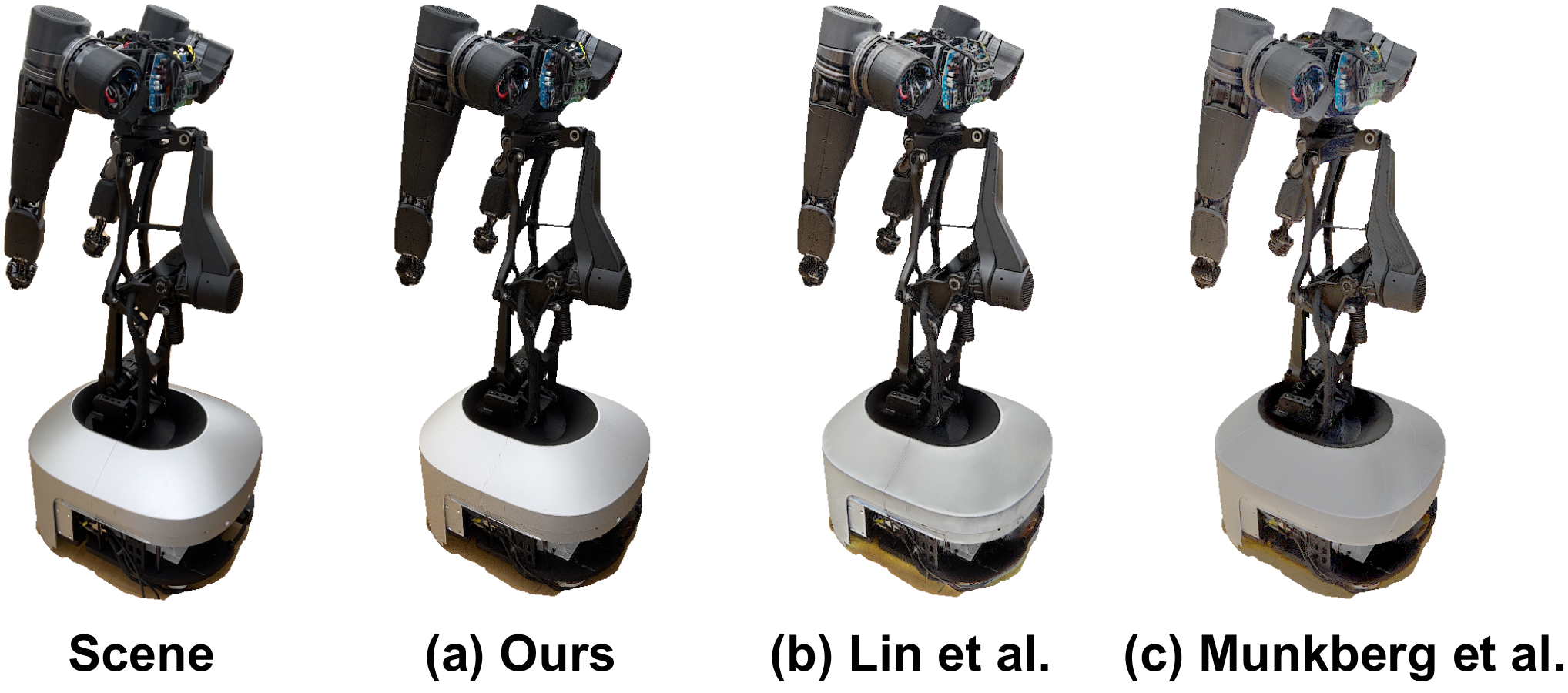}
    \vspace{-2mm}
    \caption{Qualitative comparisons of texture optimization algorithms based on differentiable rendering: (a) our method, (b)  Lin et al. \cite{9915626}, and (c) Munkberg et al. \cite{munkberg2021nvdiffrec}.}
    \label{fig:texture_ablation}
    \vspace{-4mm}
\end{figure}
          

For qualitative comparison of textured meshes, we compare our proposed method with four different methods:
1) ACMP \cite{xu2020planar} is a state-of-the-art multi-view stereo algorithm: 2) Color Map Optimization (CMO) \cite{zhou2014colormapoptimization} is a texture mapping method for RGB-D scanning: 3) Waechter et al. \cite{waechter2014TexRecon} is a large-scale texture reconstruction based on global color adjustment, 
4) NeuS \cite{wang2021neus} is the recent neural surface reconstruction based on volumetric rendering. 
We reconstruct a 3D mesh from the point cloud of ACMP with Screened Poisson Surface reconstruction \cite{kazhdan2013screened}. In the case of CMO, since explicit depth maps from low-resolution LiDAR are too noisy, we utilize depth maps rendered by a triangular mesh from our geometry reconstruction step for a fair comparison. With these rendered depth maps and corresponding images, we run a color map optimization for mapping color images onto our reconstructed mesh. To train NeuS, we follow a training process similar to that in Sec. \ref{sec:Geometry}. We then extract a mesh from the SDF network in NeuS and apply the Blender Smart UV Project tool to obtain a per-vertex UV mapping. To fill a texture for this mesh, points densely sampled from triangle mesh with interpolated UV coordinates are fed to the color network, and eventually we can obtain texture maps for the given mesh.             

In Fig. \ref{fig:arkit_texture_comparison}, we show the reconstructed texture for novel views. The visual results of our approach are visually sharper and perceptually closer to the ground truth scene. Results of NeuS and CMO exhibit blurring artifacts and color misalignment. The reconstructed mesh from ACMP is very noisy due to incorrect geometry. Here, Waechter et al. (Fig. \ref{fig:arkit_texture_comparison} (b)) is the result of classical texture reconstruction in Section \ref{sec:Texture}. This method is more photorealistic and sharper than NeuS, CMO, and ACMP. Then, after texture fine-tuning, our method avoids the seams and texture misalignment often seen with Waechter et al. Table \ref{table: metric} reports the quantitative results on two datasets. Our method outperforms other texture optimization methods on all evaluation metrics. It indicates that our method satisfies both the high fidelity of the texture and its conformity to the ground truth.

In Fig. \ref{fig:texture_ablation}, we show the results of other texture optimization methods \cite{9915626,munkberg2021nvdiffrec} based on differentiable rendering \cite{nvdiffrast}. Both methods aim to learn geometry, textures, and lighting parameters. However, they heavily rely on ground truth masks which require additional annotation costs. Without ground truth masks, we note that joint optimization over geometry and texture is unstable and often has negative effects on geometry. For a fair comparison, we therefore provide the fixed mesh topology and train only texture and lighting parameters through differentiable rendering \cite{nvdiffrast}. Since we capture objects in the wild under unknown environmental lighting conditions, it is difficult to disentangle materials and lighting via differentiable rendering. Neither method \cite{9915626,munkberg2021nvdiffrec} is able to generate realistic texture maps compared to either our method or classical texture reconstruction approaches.         


\noindent\textbf{Experimental Results on DTU dataset:} Unlike our dataset, DTU \cite{DTUdataset} is captured by specialized hardware in the controlled lab environment. Thus, this dataset provides the ground truth poses and the RGBD-aided SfM is not necessary. As shown in Table \ref{table:dtu_metric}, the surfaces obtained by COLMAP \cite{schonberger2016pixelwise} with trimming value 7 achieve high-quality reconstruction. Our method exploits MVS \cite{schonberger2016pixelwise} to supervise the training NeuS and outperforms NeuS. For the ablation study, when we only train our network with the first step (ours without 2nd step), its result is not better than our method in some scans such as scan37. In our experiment, sparse voxel octree sampling does not have a significant effect on geometry performance. Since our training process can be applied to any other 3D neural surface reconstruction methods. In the future, we will apply this training process to other recent 3D neural surface reconstruction algorithms \cite{yariv2021volsdf, NeuralWarp,yu2022monosdf}, which shows better results on DTU.  


\section{Conclusion, Limitations and Future Work}
In this paper, we have developed a practical framework to reconstruct high-quality textured meshes of 3D objects by using a smartphone. Given initial poses from ARKit, multi-view images, and low-resolution depth maps, our approach can refine initial poses and depth maps from ARKit. We adopt neural geometry representation \cite{wang2021neus} for surface reconstruction and improve its performance by modifying the training process. Finally, a differentiable rendering is designed for fine-tuning the texture map acquired from the classical texture mapping method. 
In the future, our goal is to reduce the training time which can be comparable with classical methods. Unfortunately, most neural geometry reconstruction \cite{wang2021neus,yariv2021volsdf,NeuralWarp,yu2022monosdf} requires a long training time. We would like to reduce the training time in our framework for practical purposes.    

\noindent \textbf{Acknowledgements} 
This research was partly supported by Army Cooperative Agreement W911NF2120076.

{\small
\bibliographystyle{ieee_fullname}
\bibliography{main}

\begin{thebibliography}{10}\itemsep=-1pt

\bibitem{arkit}
Apple arkit.
\newblock \url{https://developer.apple.com/documentation/arkit/}.
\newblock Accessed: 2022-11-09.

\bibitem{ARCore}
Google arcore.
\newblock \url{https://developers.google.com/ar}.
\newblock Accessed: 2022-11-09.

\bibitem{atzmon2020sal}
Matan Atzmon and Yaron Lipman.
\newblock Sal: Sign agnostic learning of shapes from raw data.
\newblock In {\em Proceedings of the IEEE/CVF Conference on Computer Vision and
  Pattern Recognition}, pages 2565--2574, 2020.

\bibitem{azinovic2022neuralrgbdsurfacerecon}
Dejan Azinovi{\'c}, Ricardo Martin-Brualla, Dan~B Goldman, Matthias
  Nie{\ss}ner, and Justus Thies.
\newblock Neural rgb-d surface reconstruction.
\newblock In {\em Proceedings of the IEEE/CVF Conference on Computer Vision and
  Pattern Recognition}, pages 6290--6301, 2022.

\bibitem{arkitscenes}
Gilad Baruch, Zhuoyuan Chen, Afshin Dehghan, Yuri Feigin, Peter Fu, Thomas
  Gebauer, Daniel Kurz, Tal Dimry, Brandon Joffe, Arik Schwartz, and Elad
  Shulman.
\newblock {ARK}itscenes: A diverse real-world dataset for 3d indoor scene
  understanding using mobile {RGB}-d data.
\newblock In {\em Thirty-fifth Conference on Neural Information Processing
  Systems Datasets and Benchmarks Track (Round 1)}, 2021.

\bibitem{bi2017patch}
Sai Bi, Nima~Khademi Kalantari, and Ravi Ramamoorthi.
\newblock Patch-based optimization for image-based texture mapping.
\newblock {\em ACM Trans. Graph.}, 36(4):106--1, 2017.

\bibitem{neumesh}
{Chong Bao and Bangbang Yang}, Zeng Junyi, Bao Hujun, Zhang Yinda, Cui
  Zhaopeng, and Zhang Guofeng.
\newblock Neumesh: Learning disentangled neural mesh-based implicit field for
  geometry and texture editing.
\newblock In {\em European Conference on Computer Vision (ECCV)}, 2022.

\bibitem{curless1996volumetric}
Brian Curless and Marc Levoy.
\newblock A volumetric method for building complex models from range images.
\newblock In {\em Proceedings of the 23rd annual conference on Computer
  graphics and interactive techniques}, pages 303--312, 1996.

\bibitem{dai2017bundlefusion}
Angela Dai, Matthias Nie{\ss}ner, Michael Zollh{\"o}fer, Shahram Izadi, and
  Christian Theobalt.
\newblock Bundlefusion: Real-time globally consistent 3d reconstruction using
  on-the-fly surface reintegration.
\newblock {\em ACM Transactions on Graphics (ToG)}, 36(4):1, 2017.

\bibitem{NeuralWarp}
Fran{\c{c}}ois Darmon, B{\'e}n{\'e}dicte Bascle, Jean-Cl{\'e}ment Devaux,
  Pascal Monasse, and Mathieu Aubry.
\newblock Improving neural implicit surfaces geometry with patch warping.
\newblock In {\em Proceedings of the IEEE/CVF Conference on Computer Vision and
  Pattern Recognition}, pages 6260--6269, 2022.

\bibitem{detone18superpoint}
Daniel DeTone, Tomasz Malisiewicz, and Andrew Rabinovich.
\newblock Superpoint: Self-supervised interest point detection and description.
\newblock In {\em CVPR Deep Learning for Visual SLAM Workshop}, 2018.

\bibitem{flint2018visual}
Alex Flint, Oleg Naroditsky, Christopher~P Broaddus, Andriy Grygorenko,
  Stergios Roumeliotis, and Oriel Bergig.
\newblock Visual-based inertial navigation, Dec.~11 2018.
\newblock US Patent 10,152,795.

\bibitem{furukawa2015multi}
Yasutaka Furukawa and Carlos Hern{\'a}ndez.
\newblock Multi-view stereo: A tutorial.
\newblock {\em Foundations and Trends{\textregistered} in Computer Graphics and
  Vision}, 9(1-2):1--148, 2015.

\bibitem{gipuma}
Silvano Galliani, Katrin Lasinger, and Konrad Schindler.
\newblock Massively parallel multiview stereopsis by surface normal diffusion.
\newblock June 2015.

\bibitem{garland1997surface}
Michael Garland and Paul~S Heckbert.
\newblock Surface simplification using quadric error metrics.
\newblock In {\em Proceedings of the 24th annual conference on Computer
  graphics and interactive techniques}, pages 209--216, 1997.

\bibitem{Goel_2022_CVPR}
Shubham Goel, Georgia Gkioxari, and Jitendra Malik.
\newblock Differentiable stereopsis: Meshes from multiple views using
  differentiable rendering.
\newblock In {\em Proceedings of the IEEE/CVF Conference on Computer Vision and
  Pattern Recognition (CVPR)}, pages 8635--8644, June 2022.

\bibitem{Kanazawa_2018ECCV2}
Shubham Goel, Angjoo Kanazawa, and Jitendra Malik.
\newblock Shape and viewpoint without keypoints.
\newblock In {\em Computer Vision – ECCV 2020: 16th European Conference,
  Glasgow, UK, August 23–28, 2020, Proceedings, Part XV}, page 88–104,
  Berlin, Heidelberg, 2020. Springer-Verlag.

\bibitem{eik}
Amos Gropp, Lior Yariv, Niv Haim, Matan Atzmon, and Yaron Lipman.
\newblock Implicit geometric regularization for learning shapes.
\newblock In {\em Proceedings of the 37th International Conference on Machine
  Learning}, ICML'20. JMLR.org, 2020.

\bibitem{manhattansdf}
Haoyu Guo, Sida Peng, Haotong Lin, Qianqian Wang, Guofeng Zhang, Hujun Bao, and
  Xiaowei Zhou.
\newblock Neural 3d scene reconstruction with the manhattan-world assumption.
\newblock In {\em Proceedings of the IEEE/CVF Conference on Computer Vision and
  Pattern Recognition}, pages 5511--5520, 2022.

\bibitem{hartley2003multiple}
Richard Hartley and Andrew Zisserman.
\newblock {\em Multiple view geometry in computer vision}.
\newblock Cambridge university press, 2003.

\bibitem{henderson2020leveraging}
Paul Henderson, Vagia Tsiminaki, and Christoph~H Lampert.
\newblock Leveraging 2d data to learn textured 3d mesh generation.
\newblock In {\em Proceedings of the IEEE/CVF Conference on Computer Vision and
  Pattern Recognition}, pages 7498--7507, 2020.

\bibitem{huang2020adversarial}
Jingwei Huang, Justus Thies, Angela Dai, Abhijit Kundu, Chiyu Jiang, Leonidas~J
  Guibas, Matthias Nie{\ss}ner, Thomas Funkhouser, et~al.
\newblock Adversarial texture optimization from rgb-d scans.
\newblock In {\em Proceedings of the IEEE/CVF Conference on Computer Vision and
  Pattern Recognition}, pages 1559--1568, 2020.

\bibitem{izadi2011kinectfusion}
Shahram Izadi, David Kim, Otmar Hilliges, David Molyneaux, Richard Newcombe,
  Pushmeet Kohli, Jamie Shotton, Steve Hodges, Dustin Freeman, Andrew Davison,
  et~al.
\newblock Kinectfusion: real-time 3d reconstruction and interaction using a
  moving depth camera.
\newblock In {\em Proceedings of the 24th annual ACM symposium on User
  interface software and technology}, pages 559--568, 2011.

\bibitem{DTUdataset}
Rasmus Jensen, Anders Dahl, George Vogiatzis, Engin Tola, and Henrik Aan{\ae}s.
\newblock Large scale multi-view stereopsis evaluation.
\newblock In {\em Proceedings of the IEEE conference on computer vision and
  pattern recognition}, pages 406--413, 2014.

\bibitem{Kanazawa_2018_ECCV}
Angjoo Kanazawa, Shubham Tulsiani, Alexei~A. Efros, and Jitendra Malik.
\newblock Learning category-specific mesh reconstruction from image
  collections.
\newblock In {\em Proceedings of the European Conference on Computer Vision
  (ECCV)}, September 2018.

\bibitem{kazhdan2006poisson}
Michael Kazhdan, Matthew Bolitho, and Hugues Hoppe.
\newblock Poisson surface reconstruction.
\newblock In {\em Proceedings of the fourth Eurographics symposium on Geometry
  processing}, volume~7, 2006.

\bibitem{kazhdan2013screened}
Michael Kazhdan and Hugues Hoppe.
\newblock Screened poisson surface reconstruction.
\newblock {\em ACM Transactions on Graphics (ToG)}, 32(3):1--13, 2013.

\bibitem{kingma2014adam}
Diederik~P Kingma and Jimmy Ba.
\newblock Adam: A method for stochastic optimization.
\newblock {\em arXiv preprint arXiv:1412.6980}, 2014.

\bibitem{Kirillov_2020_CVPR}
Alexander Kirillov, Yuxin Wu, Kaiming He, and Ross Girshick.
\newblock Pointrend: Image segmentation as rendering.
\newblock In {\em Proceedings of the IEEE/CVF Conference on Computer Vision and
  Pattern Recognition (CVPR)}, June 2020.

\bibitem{labatut2009robust}
Patrick Labatut, J-P Pons, and Renaud Keriven.
\newblock Robust and efficient surface reconstruction from range data.
\newblock In {\em Computer graphics forum}, volume~28, pages 2275--2290. Wiley
  Online Library, 2009.

\bibitem{nvdiffrast}
Samuli Laine, Janne Hellsten, Tero Karras, Yeongho Seol, Jaakko Lehtinen, and
  Timo Aila.
\newblock Modular primitives for high-performance differentiable rendering.
\newblock {\em ACM Transactions on Graphics (TOG)}, 39(6):1--14, 2020.

\bibitem{lempitsky2007seamless}
Victor Lempitsky and Denis Ivanov.
\newblock Seamless mosaicing of image-based texture maps.
\newblock In {\em 2007 IEEE conference on computer vision and pattern
  recognition}, pages 1--6. IEEE, 2007.

\bibitem{9915626}
Lixiang Lin, Jianke Zhu, and Yisu Zhang.
\newblock Multiview textured mesh recovery by differentiable rendering.
\newblock {\em IEEE Transactions on Circuits and Systems for Video Technology},
  pages 1--1, 2022.

\bibitem{sparseneus}
Xiaoxiao Long, Cheng Lin, Peng Wang, Taku Komura, and Wenping Wang.
\newblock Sparseneus: Fast generalizable neural surface reconstruction from
  sparse views.
\newblock {\em arXiv preprint arXiv:2206.05737}, 2022.

\bibitem{marchingcube}
William~E Lorensen and Harvey~E Cline.
\newblock Marching cubes: A high resolution 3d surface construction algorithm.
\newblock {\em ACM siggraph computer graphics}, 21(4):163--169, 1987.

\bibitem{lowe2004distinctive}
David~G Lowe.
\newblock Distinctive image features from scale-invariant keypoints.
\newblock {\em International journal of computer vision}, 60(2):91--110, 2004.

\bibitem{martin2021nerfinthewild}
Ricardo Martin-Brualla, Noha Radwan, Mehdi~SM Sajjadi, Jonathan~T Barron,
  Alexey Dosovitskiy, and Daniel Duckworth.
\newblock Nerf in the wild: Neural radiance fields for unconstrained photo
  collections.
\newblock In {\em Proceedings of the IEEE/CVF Conference on Computer Vision and
  Pattern Recognition}, pages 7210--7219, 2021.

\bibitem{mildenhall2021nerf}
Ben Mildenhall, Pratul~P Srinivasan, Matthew Tancik, Jonathan~T Barron, Ravi
  Ramamoorthi, and Ren Ng.
\newblock Nerf: Representing scenes as neural radiance fields for view
  synthesis.
\newblock {\em Communications of the ACM}, 65(1):99--106, 2021.

\bibitem{muller2022instant}
Thomas M{\"u}ller, Alex Evans, Christoph Schied, and Alexander Keller.
\newblock Instant neural graphics primitives with a multiresolution hash
  encoding.
\newblock {\em arXiv preprint arXiv:2201.05989}, 2022.

\bibitem{munkberg2021nvdiffrec}
Jacob Munkberg, Jon Hasselgren, Tianchang Shen, Jun Gao, Wenzheng Chen, Alex
  Evans, Thomas Mueller, and Sanja Fidler.
\newblock {Extracting Triangular 3D Models, Materials, and Lighting From
  Images}.
\newblock {\em arXiv:2111.12503}, 2021.

\bibitem{niemeyer2020differentiable}
Michael Niemeyer, Lars Mescheder, Michael Oechsle, and Andreas Geiger.
\newblock Differentiable volumetric rendering: Learning implicit 3d
  representations without 3d supervision.
\newblock In {\em Proceedings of the IEEE/CVF Conference on Computer Vision and
  Pattern Recognition}, pages 3504--3515, 2020.

\bibitem{niessner2013real}
Matthias Nie{\ss}ner, Michael Zollh{\"o}fer, Shahram Izadi, and Marc
  Stamminger.
\newblock Real-time 3d reconstruction at scale using voxel hashing.
\newblock {\em ACM Transactions on Graphics (ToG)}, 32(6):1--11, 2013.

\bibitem{oechsle2021unisurf}
Michael Oechsle, Songyou Peng, and Andreas Geiger.
\newblock Unisurf: Unifying neural implicit surfaces and radiance fields for
  multi-view reconstruction.
\newblock In {\em Proceedings of the IEEE/CVF International Conference on
  Computer Vision}, pages 5589--5599, 2021.

\bibitem{revaud2019r2d2}
Jerome Revaud, Philippe Weinzaepfel, C{\'e}sar De~Souza, Noe Pion, Gabriela
  Csurka, Yohann Cabon, and Martin Humenberger.
\newblock R2d2: repeatable and reliable detector and descriptor.
\newblock {\em arXiv preprint arXiv:1906.06195}, 2019.

\bibitem{Sarlin_2020_CVPR}
Paul-Edouard Sarlin, Daniel DeTone, Tomasz Malisiewicz, and Andrew Rabinovich.
\newblock Superglue: Learning feature matching with graph neural networks.
\newblock In {\em Proceedings of the IEEE/CVF Conference on Computer Vision and
  Pattern Recognition (CVPR)}, June 2020.

\bibitem{schonberger2016structure}
Johannes~L Schonberger and Jan-Michael Frahm.
\newblock Structure-from-motion revisited.
\newblock In {\em Proceedings of the IEEE Conference on Computer Vision and
  Pattern Recognition}, pages 4104--4113, 2016.

\bibitem{schonberger2016pixelwise}
Johannes~L Sch{\"o}nberger, Enliang Zheng, Jan-Michael Frahm, and Marc
  Pollefeys.
\newblock Pixelwise view selection for unstructured multi-view stereo.
\newblock In {\em European Conference on Computer Vision}, pages 501--518.
  Springer, 2016.

\bibitem{snavely2006photo}
Noah Snavely, Steven~M Seitz, and Richard Szeliski.
\newblock Photo tourism: exploring photo collections in 3d.
\newblock In {\em ACM siggraph 2006 papers}, pages 835--846. 2006.

\bibitem{snavely2008modeling}
Noah Snavely, Steven~M Seitz, and Richard Szeliski.
\newblock Modeling the world from internet photo collections.
\newblock {\em International journal of computer vision}, 80(2):189--210, 2008.

\bibitem{sun2022neural3dwild}
Jiaming Sun, Xi Chen, Qianqian Wang, Zhengqi Li, Hadar Averbuch-Elor, Xiaowei
  Zhou, and Noah Snavely.
\newblock Neural 3d reconstruction in the wild.
\newblock In {\em ACM SIGGRAPH 2022 Conference Proceedings}, pages 1--9, 2022.

\bibitem{10.1145/3306346.3323035}
Justus Thies, Michael Zollh\"{o}fer, and Matthias Nie\ss{}ner.
\newblock Deferred neural rendering: Image synthesis using neural textures.
\newblock {\em ACM Trans. Graph.}, 38(4), jul 2019.

\bibitem{vogiatzis2005multi}
George Vogiatzis, Philip~HS Torr, and Roberto Cipolla.
\newblock Multi-view stereo via volumetric graph-cuts.
\newblock In {\em 2005 IEEE Computer Society Conference on Computer Vision and
  Pattern Recognition (CVPR'05)}, volume~2, pages 391--398. IEEE, 2005.

\bibitem{waechter2014TexRecon}
Michael Waechter, Nils Moehrle, and Michael Goesele.
\newblock Let there be color! large-scale texturing of 3d reconstructions.
\newblock In {\em European conference on computer vision}, pages 836--850.
  Springer, 2014.

\bibitem{neuris}
Jiepeng Wang, Peng Wang, Xiaoxiao Long, Christian Theobalt, Taku Komura,
  Lingjie Liu, and Wenping Wang.
\newblock Neuris: Neural reconstruction of indoor scenes using normal priors.
\newblock {\em arXiv preprint arXiv:2206.13597}, 2022.

\bibitem{wang2021neus}
Peng Wang, Lingjie Liu, Yuan Liu, Christian Theobalt, Taku Komura, and Wenping
  Wang.
\newblock Neus: Learning neural implicit surfaces by volume rendering for
  multi-view reconstruction.
\newblock In A. Beygelzimer, Y. Dauphin, P. Liang, and J.~Wortman Vaughan,
  editors, {\em Advances in Neural Information Processing Systems}, 2021.

\bibitem{wang2004image}
Zhou Wang, Alan~C Bovik, Hamid~R Sheikh, and Eero~P Simoncelli.
\newblock Image quality assessment: from error visibility to structural
  similarity.
\newblock {\em IEEE transactions on image processing}, 13(4):600--612, 2004.

\bibitem{whelan2015elasticfusion}
Thomas Whelan, Stefan Leutenegger, Renato Salas-Moreno, Ben Glocker, and Andrew
  Davison.
\newblock Elasticfusion: Dense slam without a pose graph.
\newblock Robotics: Science and Systems, 2015.

\bibitem{widya2020stomach}
Aji~Resindra Widya, Yusuke Monno, Masatoshi Okutomi, Sho Suzuki, Takuji Gotoda,
  and Kenji Miki.
\newblock Stomach 3d reconstruction based on virtual chromoendoscopic image
  generation.
\newblock In {\em 2020 42nd Annual International Conference of the IEEE
  Engineering in Medicine \& Biology Society (EMBC)}, pages 1848--1852. IEEE,
  2020.

\bibitem{wu2013towards}
Changchang Wu.
\newblock Towards linear-time incremental structure from motion.
\newblock In {\em 2013 International Conference on 3D Vision-3DV 2013}, pages
  127--134. IEEE, 2013.

\bibitem{NeuTex}
Fanbo Xiang, Zexiang Xu, Milos Hasan, Yannick Hold-Geoffroy, Kalyan Sunkavalli,
  and Hao Su.
\newblock Neutex: Neural texture mapping for volumetric neural rendering.
\newblock In {\em Proceedings of the IEEE/CVF Conference on Computer Vision and
  Pattern Recognition (CVPR)}, pages 7119--7128, June 2021.

\bibitem{xu2022multi}
Qingshan Xu, Weihang Kong, Wenbing Tao, and Marc Pollefeys.
\newblock Multi-scale geometric consistency guided and planar prior assisted
  multi-view stereo.
\newblock {\em IEEE Transactions on Pattern Analysis and Machine Intelligence},
  2022.

\bibitem{xu2020planar}
Qingshan Xu and Wenbing Tao.
\newblock Planar prior assisted patchmatch multi-view stereo.
\newblock In {\em Proceedings of the AAAI Conference on Artificial
  Intelligence}, volume~34, pages 12516--12523, 2020.

\bibitem{yariv2021volsdf}
Lior Yariv, Jiatao Gu, Yoni Kasten, and Yaron Lipman.
\newblock Volume rendering of neural implicit surfaces.
\newblock {\em Advances in Neural Information Processing Systems},
  34:4805--4815, 2021.

\bibitem{yariv2020IDR}
Lior Yariv, Yoni Kasten, Dror Moran, Meirav Galun, Matan Atzmon, Basri Ronen,
  and Yaron Lipman.
\newblock Multiview neural surface reconstruction by disentangling geometry and
  appearance.
\newblock {\em Advances in Neural Information Processing Systems},
  33:2492--2502, 2020.

\bibitem{yu2021plenoctrees}
Alex Yu, Ruilong Li, Matthew Tancik, Hao Li, Ren Ng, and Angjoo Kanazawa.
\newblock Plenoctrees for real-time rendering of neural radiance fields.
\newblock In {\em Proceedings of the IEEE/CVF International Conference on
  Computer Vision}, pages 5752--5761, 2021.

\bibitem{yu2022monosdf}
Zehao Yu, Songyou Peng, Michael Niemeyer, Torsten Sattler, and Andreas Geiger.
\newblock Monosdf: Exploring monocular geometric cues for neural implicit
  surface reconstruction.
\newblock {\em arXiv preprint arXiv:2206.00665}, 2022.

\bibitem{mvsdf}
Jingyang Zhang, Yao Yao, and Long Quan.
\newblock Learning signed distance field for multi-view surface reconstruction.
\newblock In {\em Proceedings of the IEEE/CVF International Conference on
  Computer Vision}, pages 6525--6534, 2021.

\bibitem{zhang2020nerf++}
Kai Zhang, Gernot Riegler, Noah Snavely, and Vladlen Koltun.
\newblock Nerf++: Analyzing and improving neural radiance fields.
\newblock {\em arXiv preprint arXiv:2010.07492}, 2020.

\bibitem{zhou2014colormapoptimization}
Qian-Yi Zhou and Vladlen Koltun.
\newblock Color map optimization for 3d reconstruction with consumer depth
  cameras.
\newblock {\em ACM Transactions on Graphics (ToG)}, 33(4):1--10, 2014.

\bibitem{zollhofer2018state}
Michael Zollh{\"o}fer, Patrick Stotko, Andreas G{\"o}rlitz, Christian Theobalt,
  Matthias Nie{\ss}ner, Reinhard Klein, and Andreas Kolb.
\newblock State of the art on 3d reconstruction with rgb-d cameras.
\newblock In {\em Computer graphics forum}, volume~37, pages 625--652. Wiley
  Online Library, 2018.

\end{thebibliography}
}

\end{document}